# Dataset for eye-tracking tasks

Rakhmatulin Ildar, PhD
*South Ural State University, Department of Power Plants Networks and Systems, Chelyabinsk city, Russia, 454080*
ildar.o2010@yandex.ru

**Abstract**
In recent years many different deep neural networks were developed, but due to a large number of layers in deep networks, their training requires a long time and a large number of datasets. Today is popular to use trained deep neural networks for various tasks, even for simple ones in which such deep networks are not required. The well-known deep networks such as YoloV3, SSD, etc. are intended for tracking and monitoring various objects, therefore their weights are heavy and the overall accuracy for a specific task is low. Eye-tracking tasks need to detect only one object - an iris in a given area. Therefore, it is logical to use a neural network only for this task. But the problem is the lack of suitable datasets for training the model. In the manuscript, we presented a dataset that is suitable for training custom models of convolutional neural networks for eye-tracking tasks. Using data set data, each user can independently pre-train the convolutional neural network models for eye-tracking tasks. This dataset contains annotated 10,000 eye images in an extension of 416 by 416 pixels. The table with annotation information shows the coordinates and radius of the eye for each image. This manuscript can be considered as a guide for the preparation of datasets for eye-tracking devices.
**Program by python in Github:** https://github.com/Ildaron/5.eye_tracking_with_CNN
**Dataset in Kaggle:** www.kaggle.com/ildaron/dataset-eyetracking

**Data value**
We provided a fully labeled dataset with eye position in an image with a resolution of 416 by 416 pixels. The dataset can be used to develop the development of convolutional neural networks for the detection, segmentation, and classification of the position of the iris. Using data from a data set, each user can independently train a neural network using a small set of personal data to search for a specific (user) type of iris.

**1. Introduction**
Today Eye-tracking is used to support multimedia learning, help in browsing the web, and is widely used in real-time graphics systems, which is especially popular in video games. The main problem of modern Eye-tracking systems is their high price. Equipment with accuracy of 0.5 ° has prices from several thousand dollars. The most common distance eye-trackers use the corneal reflection method (CR). The eyes are exposed to direct invisible infrared (IR) light, which results in the reflection in the cornea. The physiology of this process is described in detail in the manuscript [1]. The next research that uses well-known deep networks with trained weights [2,3].
The neural network model must have a high accuracy of iris recognition. The color of the eyes of each person is unique, as a result of which the neural network should focus on the characteristics that are not directly related to color since it is not possible to train the network for all possible colors. In this paper, a dataset is presented in the resolution of the allowing neural network to identify useful signs for recognizing the position of the pupil. The dataset is designed to search for signs between the iris and sclera.
Figure 1 shows the process of selecting an image extension to create a dataset.

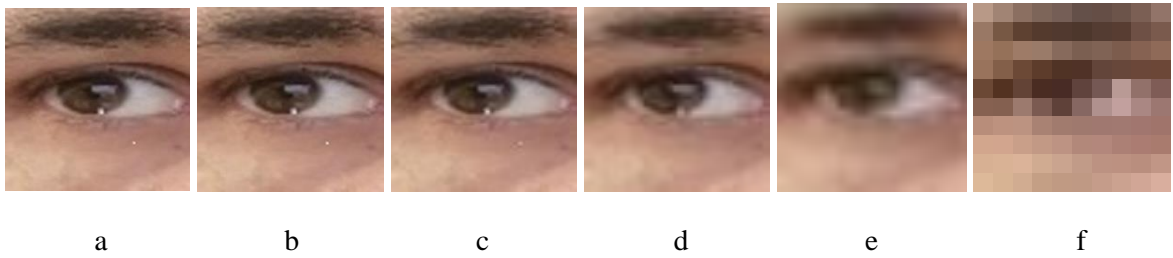

a  b  c  d  e  f

Fig.1. Examples of images with different resolutions: a - 416x416 pixels, b - 200x200 pixels, c - 100x100 pixels, d - 50x50 pixels, e - 25x25 pixels, f - 10x10 pixels

Visually, it is difficult to notice the difference between an image with a resolution of 416 by 416 pixels and 50 by 50 pixels. But this dataset is designed to determine the following features, Fig. 2.

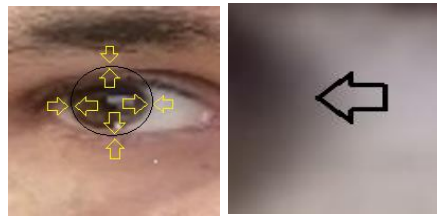

Fig.2. Features for a neural network

As a result, to more accurately determine the boundary between the iris and sclera we decided to use an image with a resolution of 416x416 pixels.
To see how this image will look between the layers, we created an ultraprecise neural network and saved various variants of images between the layers, Fig. 3.

```
Model: "sequential_1"
_________________________________________________________________
Layer (type)                 Output Shape              Param #
=================================================================
conv2d_1 (Conv2D)            (None, 414, 414, 32)      320
_________________________________________________________________
max_pooling2d_1 (MaxPooling2 (None, 207, 207, 32)      0
_________________________________________________________________
conv2d_2 (Conv2D)            (None, 205, 205, 32)      9248
_________________________________________________________________
max_pooling2d_2 (MaxPooling2 (None, 102, 102, 32)      0
_________________________________________________________________
conv2d_3 (Conv2D)            (None, 100, 100, 32)      9248
_________________________________________________________________
max_pooling2d_3 (MaxPooling2 (None, 50, 50, 32)        0
_________________________________________________________________
conv2d_4 (Conv2D)            (None, 48, 48, 32)        9248
_________________________________________________________________
max_pooling2d_4 (MaxPooling2 (None, 24, 24, 32)        0
_________________________________________________________________
conv2d_5 (Conv2D)            (None, 22, 22, 32)        9248
_________________________________________________________________
max_pooling2d_5 (MaxPooling2 (None, 11, 11, 32)        0
_________________________________________________________________
conv2d_6 (Conv2D)            (None, 9, 9, 32)          9248
_________________________________________________________________
flatten_1 (Flatten)          (None, 2592)              0
_________________________________________________________________
dense_1 (Dense)              (None, 600)               1555800
_________________________________________________________________
dropout_1 (Dropout)          (None, 600)               0
_________________________________________________________________
dense_2 (Dense)              (None, 100)               60100
_________________________________________________________________
dropout_2 (Dropout)          (None, 100)               0
_________________________________________________________________
dense_3 (Dense)              (None, 3)                 303
=================================================================
Total params: 1,662,763
Trainable params: 1,662,763
Non-trainable params: 0
_________________________________________________________________
```

Fig. 3. Neural Convolution Network Diagram

To see how this image will look between the layers, we created the next convolutional neural network, Fig. 3.

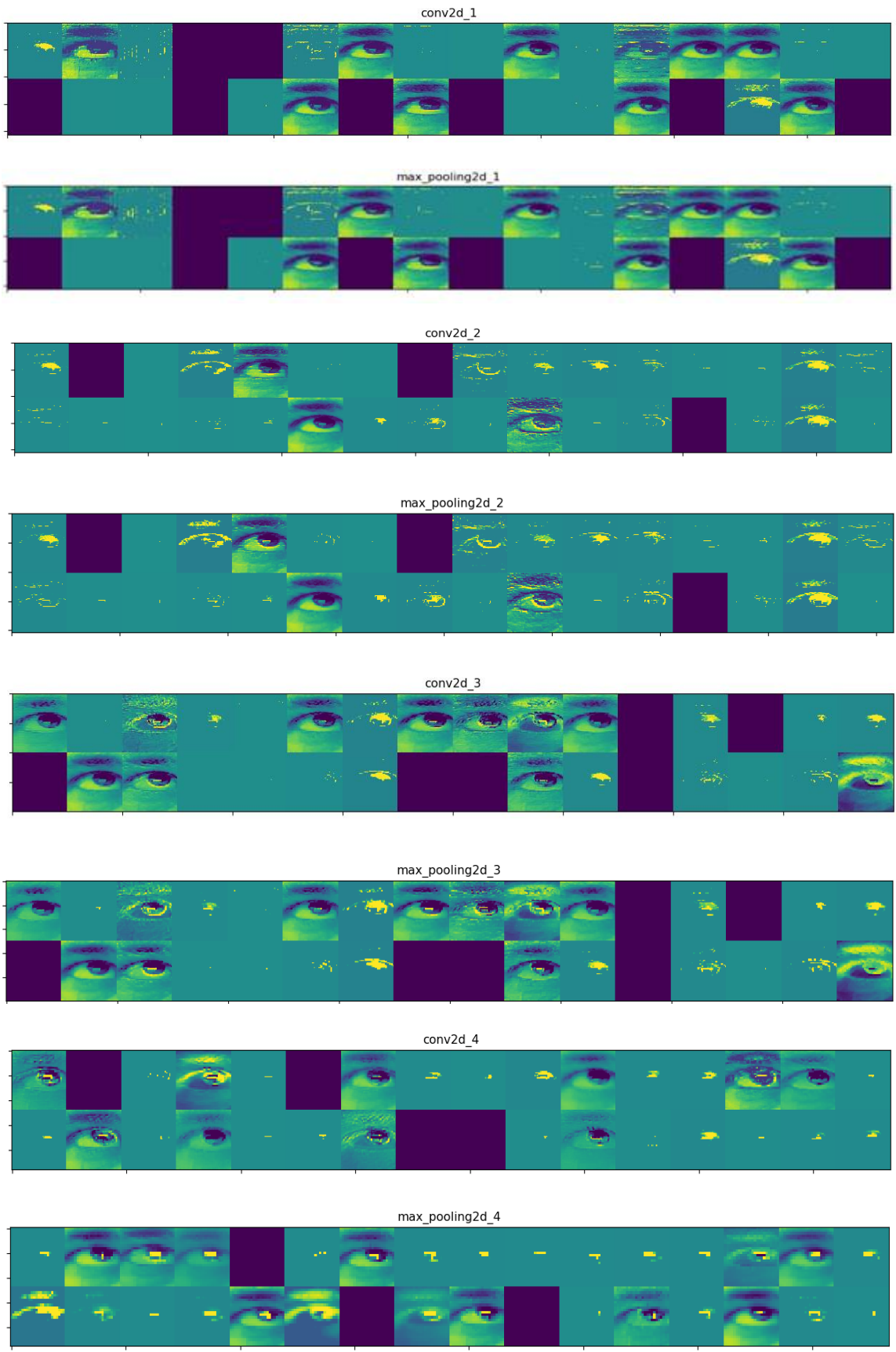

Fig. 4. Visualization of the input image on different layers of the neural network model

Visual analysis of the images shows that already on the 4th layer of the convolutional neural network, we can get the necessary signs to determine the position of the eye. Therefore, a dataset with sufficient eye area expansion is needed. Consistent with the overview of Eye-tracking, Datasets by Winkler, S., et al. [4], and an analysis of the available datasets, we considered the following sources:

- Columbia Gaze Data Set. The data set consists of 5,880 images of 56 people over varying gaze direction and head poses. For each subject, there are 5 head poses and 21 gaze directions per head pose(http://www.cs.columbia.edu/CAVE/databases/columbia gaze/). Rajeev, R. et al. used this dataset to train the neural networks [5];

- Openeds facebook dataset. Semantic segmentation data set collected with 152 participants of 12,759 images with annotations at a resolution of 400×640. Challenge participation deadline: September 15, 2019. But Dataset is still available on request (https://research.fb.com/programs/openeds-challenge/). Aayushy, C. et al. used this dataset for research [6];

- MPIIGaze dataset. This data set consists of images taken in everyday conditions using the laptop's built-in webcams, in which 15 people participate. MPIIGaze dataset that contains 213,659 images [7];

- Kaggle dataset. Competitions are held periodically, participation in which open access to the dataset. For example the next dataset https://www.kaggle.com/c/gl-eye-tracking;

To create our dataset, we selected the following face image collection (https://www.kaggle.com/4quant/eye-gaze) implemented for research by Shrivastava. A., et al. [8].

Initially, the images, image resolution of 1280 by 720 pixels, presented in the dataset are as follows, fig. 5.

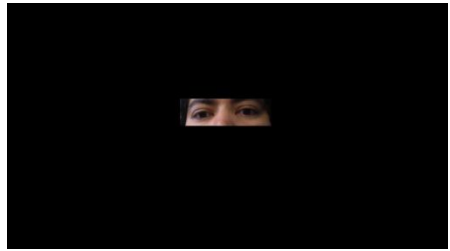

Fig. 5. Example from the image collection

To determine the eye area in the images, we used the dlib library. The landmark detection algorithm proposed by dlib is an implementation of the Regression Tree Ensemble (ERT), introduced in 2014 by Casemi and Sullivan. This method uses a simple and quick function to directly estimate the location of a landmark. These estimated positions are subsequently refined using an iterative process performed by a cascade of regressors. Regressors make a new estimate from the previous one, trying to reduce the error of alignment of the estimated points at each iteration. At the first stage, the dlib.get_frontal_face_detector () function determines the face contour. Next, using the dlib.shape_predictor command ("shape_predictor_68_face_landmarks.dat") we define facial features. Where shape_predictor_68_face_landmarks.dat is a trained model for 68 landmarks. We only take points 36 to 41 and add 20 millimeters each point to expand the range, and then using the OpenCV ROI we limit the area with the eye in the video stream, fig. 6.

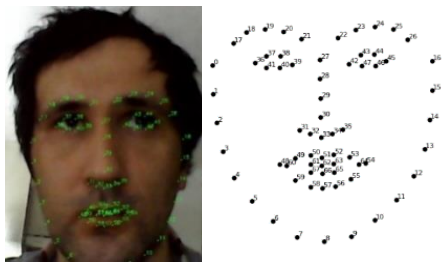

аю.

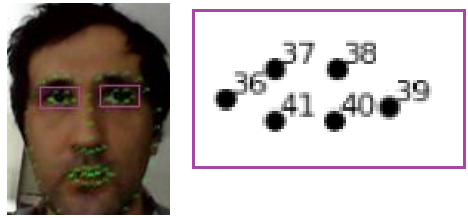

b

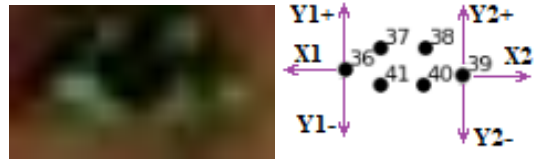

c

Fig.6. a - face detection, b – new image with points 36 to 41, c – expand new image with points 36 to 41

This operation gives us an image with pixels 77 by 55 pixels, we increase by 416 and 416 pixels, as a result, we get the following image, Fig. 7.

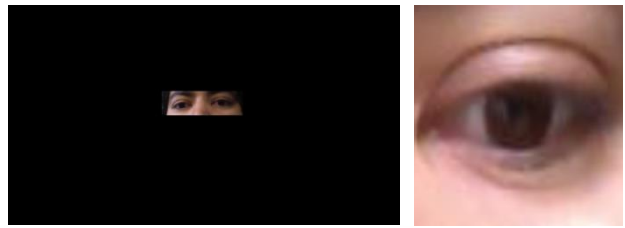

a                                    b

Fig. 7. Images from the dataset: a - before converted, b - after converted

Next, the most time-consuming part of the research, finding the iris on the processed images were realized. Taking into account that the eye area was selected using the dlib library, specifically for this set of images, we studied in detail the ratio of the pupil size to the size of the eye area, which ultimately amounted to 14%. Next, we wrote the program that, by enumerating various values of the threshold function - THRESH_BINARY, selected an image in which the iris would have a size of 14% relative to the image with the eye area. This code is presented in GitHub in the file 1.Convert_the_eye_Thershold.py. The algorithm for obtaining the dataset is shown in Fig. 5.

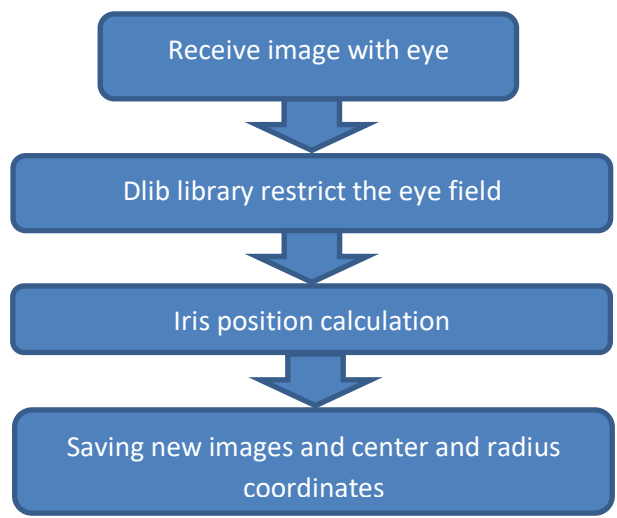

Fig.8. Algorithm for obtaining the dataset

The implementation of this algorithm for image acquisition, for clarity, is presented in the images, fig.8

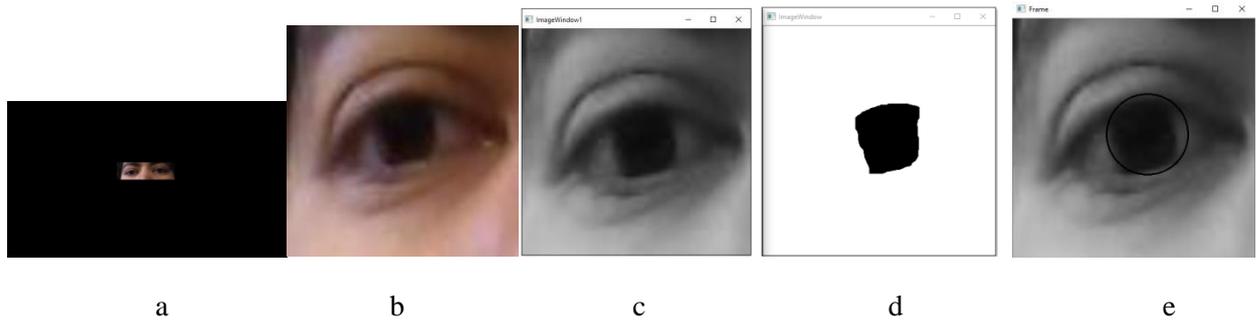

    a        b        c        d        e

Fig.8. The process of obtaining images for dataset: a – initial image, b – eye filed, c –after filters, d – iris mask, e – final image with circle around iris

Figure 9 shows the data storage format for the iris position.

| | radius | X | Y | image |
|---|---|---|---|---|
| 2 | 42.66829 | 257.219 | 141.4265 | 0 |
| 3 | 55.40091 | 221 | 184.5 | 1 |
| 4 | 26.53781 | 255 | 145.5 | 2 |
| 5 | 26.53781 | 255 | 145.5 | 3 |
| 6 | 28.29321 | 251.5 | 134.5 | 4 |
| 7 | 37.78191 | 268.0517 | 144.3734 | 5 |
| 8 | 37.78191 | 268.0517 | 144.3734 | 6 |
| 9 | 37.78191 | 268.0517 | 144.3734 | 7 |
| 10 | 37.78191 | 268.0517 | 144.3734 | 8 |
| 11 | 40.92177 | 255.2645 | 154.3348 | 9 |
| 12 | 61.57323 | 247 | 146.5 | 10 |
| 13 | 60.28484 | 236 | 215.5 | 11 |
| 14 | 57.13107 | 232.2973 | 138.5754 | 12 |

Fig.9. format for the iris position

**Conclusion and discussion**
As a result, 10,000 images were obtained with the coordinates of the center of the pupil and the radius. For annotating images, a set of images with a resolution of 1280 by 720 pixels was used. To convert the images, the dlib library allocated the eye region with a resolution of 77 by 55 pixels, later the OpenCV library increased the resolution to a scale of 416 by 416 pixels. After we created the program with an experimentally obtained equation that allows identifying the iris in the image. This dataset is intended for pre-training models of convolutional neural networks for the eye-tracking tasks.
This dataset was tested on its own model of the convolutional neural network for training the initial layers of the neural network model. To train the last layers, a personal dataset of 1000 photos was used. As a result, the tracking error was three degrees. Given that the tracking was carried out on a web camera is a good result.

The author certifies that he has NO affiliations with or involvement in any organization or entity with any financial interest (such as honoraria; educational grants; participation in speakers' bureaus; membership, employment, consultancies, stock ownership, or other equity interest; and expert testimony or patent-licensing arrangements), or non-financial interest (such as personal or professional relationships, affiliations, knowledge or beliefs) in the subject matter or materials discussed in this manuscript.

**References**

1. Hari, S. (2012). Human Eye Tracking and Related Issues: A Review. International Journal of Scientific and Research Publications, 2, Issue 9, 1-9, ISSN 2250-3153

2. Wibirama, S., Nugroho, H., & Hamamoto, K. (2017). Evaluating 3D gaze tracking in virtual space: A computer graphics approach. Entertainment Computing, 21, 11-17

3. Skodras, E., Kanas, V., & Fakotakis, N. (2015). On visual gaze tracking based on a single low cost camera. Signal Processing: Image Communication, 36, 29-42

4. Winkler, S., Subramanian, R. (2013). Overview of Eye tracking Datasets. Conference: Quality of Multimedia Experience (QoMEX), DOI: 10.1109/QoMEX.2013.6603239

5. Rajeev, R., & Shalini, D. (2018). Light-weight head pose invariant gaze tracking , IEEE Conference on Computer Vision and Pattern Recognition Workshop, arXiv:1804.08572 [cs.CV]

6. Rayushy, C., & Rakshit, K. (2019). Ritnet: Real-time semantic segmentation of the eye for gaze tracking, arXiv:1910.00694v1 [cs.CV]

7. Xucong, Z., Yusuke, S., & Mario, F. (2015). Appearance-based gaze estimation in the wild. Computer Vision and Pattern Recognition, 23,456–459. doi:10.1016/j.procs.2016.07.013

8. Shrivastava, A., Pfister, T., Tuzel, O., & Susskind, J. (2016). Learning from Simulated and Unsupervised Images through Adversarial Training. Conference: Quality of Multimedia Experience (QoMEX), 2013 Fifth International Workshop on, arXiv:1612.07828v1 [cs.CV]